\documentclass[a4paper, 10pt, conference]{ieeeconf}      % Use this line for a4 paper

\IEEEoverridecommandlockouts                              % This command is only needed if 
\usepackage{amsmath} % assumes amsmath package installed
\usepackage{amssymb}  % assumes amsmath package installed
\usepackage{url}
\usepackage{doi}
\usepackage{graphicx}

\title{\LARGE \bf
Bird-Inspired Spatial Flapping Wing Mechanism via Coupled Linkages with Single Actuator*
}
\author{Daniel Huczala$^{1}$, Sun-Pill Jung$^{2}$, and Frank C. Park$^{1}$% <-this % stops a space
\thanks{*This work was supported 
in part by IITP-MSIT under Grant 2022-220480 and Grant RS-2022-II220480 (Training and Inference Methods for Goal Oriented AI Agents),
in part by IITP-MSIT under Grant RS-2024-00436680 (Collaborative Research Projects with Microsoft Research) through Global Research Support Program in the Digital Field program,
in part by MOTIR under Grant RS-2025-25462891,
in part by MOTIR under Grant RS-2025-25460896,
in part by SNU-IPAI,
in part by SNU-AIIS,
in part by SNU-IAMD,
in part by SNU BK21+ Program in Mechanical Engineering, 
in part by SNU Institute for Engineering Research,
in part by Hyundai Motor Company and Kia.}% <-this % stops a space
\thanks{$^{1}$Daniel Huczala and Frank C. Park are with the Robotics Laboratory, 
Seoul National University, 1 Gwanak-ro, Gwanak-gu, Seoul 08826, Korea
        {\tt\small daniel.huczala@snu.ac.kr}}%
\thanks{$^{2}$Sun-Pill Jung is with the Biorobotics Laboratory,
Seoul National University, 1 Gwanak-ro, Gwanak-gu, Seoul 08826, Korea.}%
}

\usepackage{xcolor}
\usepackage{hyperref}
\usepackage{fancyhdr}

\fancypagestyle{firstpage}{
  \fancyhf{}  % clear all header/footer fields
  \fancyhead[C]{\scriptsize
    \vspace{-120pt}
    \parbox{\textwidth}{
      \textcopyright~2026 IEEE. Personal use of this material is permitted.
        Permission from IEEE must be obtained for all other uses, in any current
        or future media, including reprinting/republishing this material for
        advertising or promotional purposes, creating new collective works, for
        resale or redistribution to servers or lists, or reuse of any copyrighted
        component of this work in other works.\\
        This is the accepted manuscript. The version of record is: D. Huczala, S.-P. Jung, and F. C. Park, ``Bird-Inspired
        Spatial Flapping Wing Mechanism via Coupled Linkages with Single
        Actuator,'' in \textit{2026 23rd International Conference on Ubiquitous
        Robots (UR)}, IEEE, July 2026, pp. 102--106,
        doi: doi: \textcolor{magenta}{\href{https://doi.org/10.1109/UR69537.2026.11626707}{10.1109/UR69537.2026.11626707}}.
        \vspace{4pt}
    }}    
  
}

\begin{document}
\maketitle
\thispagestyle{firstpage}

% \maketitle
% \thispagestyle{empty}
% \pagestyle{empty}

%%%%%%%%%%%%%%%%%%%%%%%%%%%%%%%%%%%%%%%%%%%%%%%%%%%%%%%%%%%%%%%%%%%%%%%%%%%%%%%%
\begin{abstract}
Spatial single-loop mechanisms such as Bennett linkages offer a unique combination of one-degree-of-freedom actuation and nontrivial spatial trajectories, making them attractive for lightweight bio-inspired robotic design. However, existing synthesis methods rely on dual quaternion algebra and rational motion interpolation and factorization, placing task-based linkage design outside the reach of most engineers, a gap addressed by the Rational Linkages library recently developed by the authors. This paper presents a bird-inspired flapping-wing mechanism built from two coupled spatial four-bars, driven by a single motor. One linkage is actuated to generate the desired spatial sweeping stroke, while the serially coupled linkage remains unactuated and passively switches between extended and folded wing configurations over the stroke cycle. Additionally, we introduce a simplified kinematic methodology for constructing Bennett linkages from quadrilaterals that contain a desired surface area and further leverage mechanically induced passive state switching. This architecture realizes a coordinated sweep-and-fold wing motion with a single actuation input, reducing weight and control complexity. A 3D-printed prototype is assembled and tested, demonstrating the intended spatial stroke and passive folding behavior.

% Old version:
%The spatial single-loop mechanisms benefit from one degree of freedom property while keeping the ability to perform complicated trajectories and exhibit self-folding configurations. Together with a lightweight design, they can suit the needs of bio-inspired robotics applications. However, although they appear simple and elegant, the geometric task-based synthesis is rather complicated, requiring deep mathematical knowledge, and therefore often avoided in engineering tasks. To prove their potential, two coupled spatial four-bar mechanisms (Bennett linkages) in the design of the robotic flapping wing are presented. We introduce a simplified kinematic methodology for constructing Bennett linkages from quadrilaterals and additionally exploit the embodied behavior of the design. This results in a flapping wing mechanism with spatial sweep and self-folding, using only one actuator. The results are supported by laboratory experiment with an assembled prototype.

\end{abstract}

%%%%%%%%%%%%%%%%%%%%%%%%%%%%%%%%%%%%%%%%%%%%%%%%%%%%%%%%%%%%%%%%%%%%%%%%%%%%%%%%
\newcommand{\qi}{\mathbf{i}}
\newcommand{\qj}{\mathbf{j}}
\newcommand{\qk}{\mathbf{k}}

% \linenumbers

\section{INTRODUCTION}

The flapping behavior observed in birds provides powerful principles for extending the mobility and agility of robots, and a wide range of bio-inspired and biomimetic bird robots has therefore been developed. However, reproducing the high power density and compliance of biological musculoskeletal systems with artificial actuators of comparable scale and weight remains difficult. As a result, many systems rely either on rather simple design with many compromises or on complex mechanical structures that approximate the spatial motions. A key characteristic of bird-like flapping is that it is not merely a reciprocating motion, but a coupled behavior combining spatial flapping/sweeping and stroke-dependent wing extension/folding. In general, the wing is extended during downstroke to maintain a large effective aerodynamic area, whereas during upstroke it is folded to reduce air resistance. Therefore, an engineering mechanism intended to realize bird-like flapping should implement these two motion components in a coordinated manner.

Many systems realize parts of this requirement, but they show different limitations depending on the actuation architecture. First, designs using a single actuator \cite{Ajanic2020bioinsp, Chang2020softbio, O_Connor2025thesynt} are advantageous in terms of reduced mechanism complexity and lower system mass, but they often fail to reproduce the key coupled features of bird-like flapping, namely spatial flapping together with stroke-dependent wing extension/folding. For example, in \cite{Ajanic2020bioinsp, Chang2020softbio}, changes in wing shape or projected area are achieved mechanically, but thrust is generated by separate propellers, so the resulting systems are structurally different from bird-like fliers in which the wings themselves directly generate propulsion and maneuvering forces. Also, an elegant solution was proposed in \cite{O_Connor2025thesynt} using purely spherical mechanism structures inspired by hummingbird movement, but there the flying would be challenging due to the limits of the spherical design itself.

By contrast, multi-actuator avian-wing robotic systems such as \cite{Ajanic2022robotic, Chen2025flappin} improve bird-like motion fidelity by explicitly combining spatial flapping with wing folding. In the design by Ajanic et al. \cite{Ajanic2022robotic}, a flapping mechanism is combined with whole-wing rotation for pitch adjustment, and an additional actuator is used to fold the wing during upstroke in order to reduce air resistance. A similar approach was adopted by Chen et al. \cite{Chen2025flappin} in the RoboFalcon project, where the number of actuators was reduced to two while still assigning separate roles to wing folding and spatial sweep generation. However, such approaches generally require sophisticated multi-linkage architectures to drive motion components.
%, such as flapping, pitch adjustment, and folding, in a more independent manner. 
As a consequence, they incur greater complexity of the mechanism, larger system size and mass, and a greater control burden. In summary, prior work exhibits a clear trade-off: single-actuator designs tend to sacrifice bird-like fidelity, whereas multi-actuator designs improve motion fidelity at the cost of complexity and weight.

In this study, we introduce an approach that 
%reduces the number of actuated degrees of freedom while achieving both the desired spatial flapping motion and wing extension/folding 
resolves this trade-off. Unlike prior single-actuator designs \cite{Ajanic2020bioinsp, Chang2020softbio, O_Connor2025thesynt}, which sacrifice spatial flapping or stroke-dependent wing folding, our design achieves both without the mechanical complexity of multi-actuator systems,
by serially coupling two Bennett mechanisms, one actuated and the other passively driven. A Bennett linkage is a spatial four-revolute closed chain (4R) with a single degree of freedom (1-DoF), while still being capable of generating nontrivial spatial trajectories.

The key concept is to use one actuated Bennett linkage to generate the spatial flapping motion, and a second coupled Bennett linkage to provide the wing extension/folding transition through a passive degree of freedom. In particular, the Bennett mechanism responsible for folding is designed with partial mobility restriction so that the structure does not rotate continuously, but instead switches between extended and folded wing configurations according to the stroke phase (downstroke/upstroke). This embodied behavior can be interpreted as a mechanically induced passive state-switching response arising from physical interaction, where part of the motion is produced by the body morphology itself in response to external inputs rather than explicit neural control \cite{Husbands2021recenta}. In this way, it allows synchronized spatial flapping of both wings together with stroke-dependent wing-state transition using only a single actuator.

The great advantage of Bennett motion is its rational parameterization, i.e., all mechanism parts follow continuous rational trajectories and therefore the direct and inverse kinematics problems are comparatively tractable \cite{Huczala2025directk}. The motion can be generated for given poses \cite{Brunnthaler2005new}, and the mechanism synthesis problem, i.e., obtaining revolute axes from the motion, can be addressed using the rational motion factorization method \cite{Li2019factori}. These properties provide a useful theoretical basis for the synthesis and analysis of the proposed spatial flapping mechanism.

This paper is organized as follows. The mathematical background is introduced in Sect.~\ref{sec:math}. The kinematics methodology and modeling are presented in two parts of Sect.~\ref{sec:methods}, showing first the simplified spatial linkage design from a quadrilateral, then a design strategy for the spatial flap and coupling of the two linkages. Finally, the prototype is presented in Sect.~\ref{sec:Prototype}. The results are then summarized, and future work is discussed in Sect.~\ref{sec:conclusion}.

\begin{figure}[tb]
  \centering
  \includegraphics[width=0.6\columnwidth]{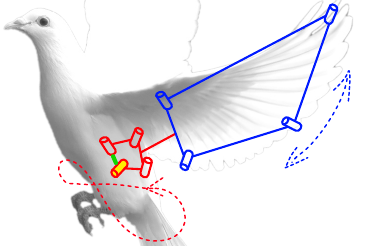}
  \caption{Schematic drawing of the proposed wing design. Red: linkage for spatial flapping; yellow: actuated joint; green: base link (bird body); blue: embodied (passive) linkage providing trust in the forward stroke and automatically folding itself during backward motion.}
  \label{fig:bird}
\end{figure}

\section{PRELIMINARIES}\label{sec:math}

The Bennett linkage is an overconstrained mechanism \cite{Perez2003dimensi}. To allow for 1-DoF mobility, the linkage must satisfy geometric constraints known as the Bennett condition
\begin{gather*}
    d_0 = d_1 = d_2 = d_3 = 0, \\
    a_0 = a_2, \quad a_1 = a_3, \\
    \alpha_0 = \alpha_2, \quad \alpha_1 = \alpha_3,
\end{gather*}
\begin{equation}
    \label{eq:bennett_cond}
    \frac{\sin \alpha_0}{\sin \alpha_1} = \frac{a_0}{a_1},
\end{equation}
where $a_i$, $\alpha_i$, and $d_i$ are Denavit-Hartenberg (DH) parameters of the $i$-th joint, $i = 0\dots3$.

This paper utilizes, especially in Sect.~\ref{sec:flapping_bennett}, results from algebraic geometry, namely the rational factorization of motion polynomials \cite{Li2019factori}. This powerful algorithm decomposes rational motions into linear factors that correspond to simple rotations or translations, that is, they can be realized by revolute or prismatic joints, and when coupled in a chain, they perform the given motion. Related methodologies also exist that allow for motion generation from input poses or points \cite{Huczala2026ark}; these were implemented by the author of this paper in the Python library Rational Linkages \cite{Huczala2024ark}. Without going into much detail, we refer to the aforementioned works for further details and implementation. However, the reader should be aware of the following definitions.

The methodology above relies on Eduard Study's rigid-body kinematics model, which introduces Study parameters related to the dual quaternion space. Dual quaternions are a vector alternative to transformation matrices: both can represent displacements of the special Euclidean group SE(3), and a mapping between the two exists~\cite[Chap.~2.8]{Stigger2019}. A dual quaternion $h \in \mathbb{DH}$ (Dual Hamiltonian) is a member of the projective seven-dimensional space $\mathbb{PR}^7$
\begin{equation}
        h = p + \varepsilon q
        = p_0 + p_1 \mathbf{i} + p_2 \mathbf{j} + p_3 \mathbf{k} 
        + \varepsilon (q_0 + q_1 \mathbf{i} + q_2 \mathbf{j} + q_3 \mathbf{k}),
\end{equation}
where $p$ and $q$ are quaternions $p, q \in \mathbb{H}$, and $\varepsilon$ is the dual unit satisfying $\varepsilon^2 = 0$. The vector representation of $h = [p_0, p_1, p_2, p_3, q_0, q_1, q_2, q_3]$ is known as the Study parameters. 

Analogously to transformation matrices -- where the rotation submatrix must have determinant equal to 1 to represent a valid rigid-body displacement -- dual quaternions must satisfy the so-called Study condition: the point in dual quaternion space representing a pose must lie on the six-dimensional Study quadric, given by
\begin{equation}
    S = p_0q_0 + p_1q_1 + p_2q_2 + p_3q_3 = 0.
    \label{studyquadric}
\end{equation}
Dual quaternions allow efficient handling of motion polynomials. For example, a rational one-parametric motion corresponds to a curve of the form $C(t) \subset S$ and can be expressed as a dual quaternion polynomial $C = p + \varepsilon q \in \mathbb{DH}[t]$.

\section{METHODOLOGY AND DESIGN}\label{sec:methods}

The wing design will consist of two 4R spatial linkages, both of which are 1-DoF mechanisms, as shown in Fig.~\ref{fig:bird}. The smaller linkage, referred to as the Stroke-linkage, will be actuated to provide the spatial flapping motion, while the larger coupled linkage, referred to as the Folding-linkage, will passively extend and fold the wing in response to the motion of the Stroke-linkage. We advise the reader to see the supplementary video (also available online in \cite{zenodo_suppl}) before going in details of the design methodology.

\subsection{Quadrilateral Design of Folding-linkage}\label{sec:quad_linkage}

During the downstroke motion, to provide thrust, maximizing the surface area acting against air resistance and gravity is required. During the upstroke, however, the wing surface area must be minimized~\cite{Ajanic2022robotic}. Single-loop linkages often exhibit expanded and folded configurations that perfectly satisfy these requirements. For example, the Bennett linkage reaches a folded configuration when the joint axes share a common transversal line.

In this section, we introduce a methodology to construct a four-bar linkage beginning from two defined quadrilaterals (this differs from the Bennett linkage constructed through a set of three poses in Sect.~\ref{sec:flapping_bennett}). We use a symmetric (rectangular) planar four-bar mechanism as an intermediate step. The rectangular shape ensures that the Bennett condition~\eqref{eq:bennett_cond} is satisfied.

\begin{figure}[t]
  \centering
  \includegraphics[width=0.98\columnwidth]{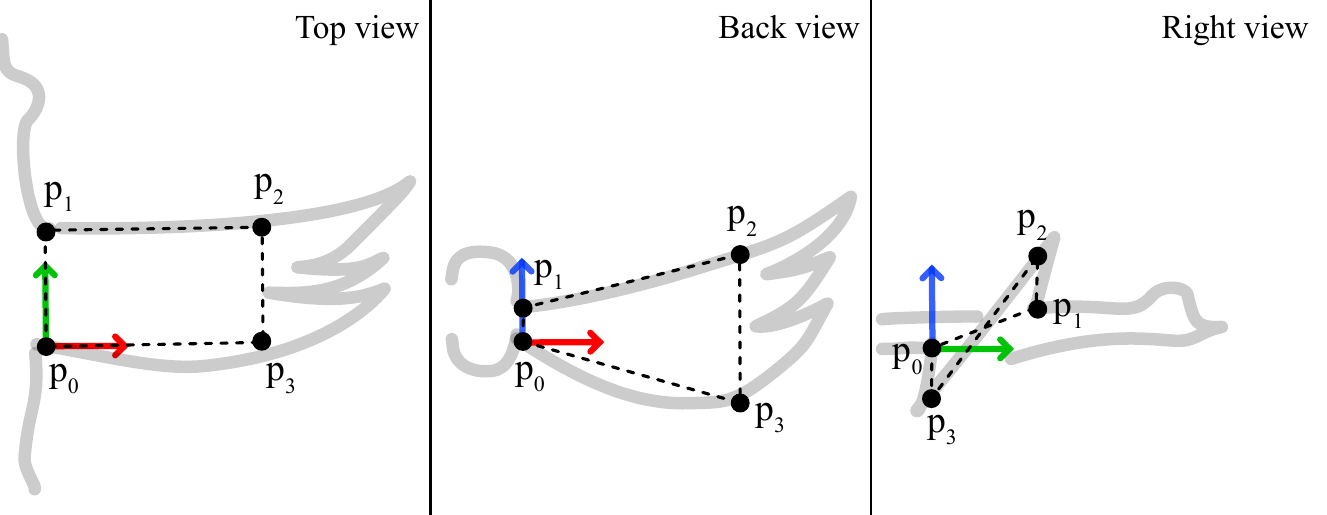}
  \caption{Quadrilateral $p_{0\dots3}$ in expanded configuration during downstroke.}
  \label{fig:quad_expanded}
\end{figure}

\begin{figure}[t]
  \centering
  \includegraphics[width=\columnwidth]{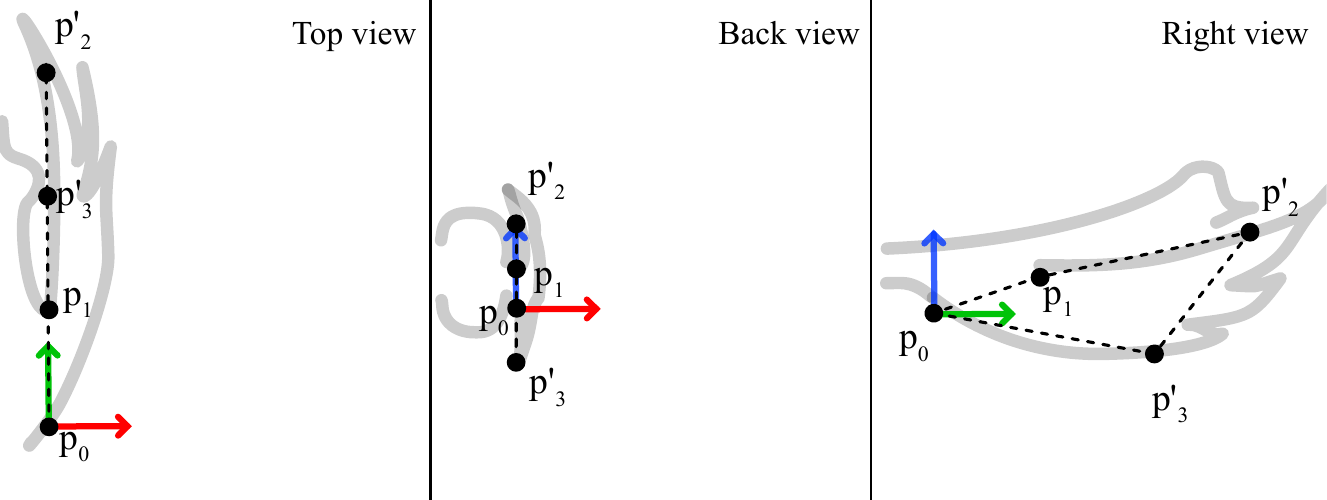}
  \caption{Quadrilateral $p_0, p_1, p'_2, p'_3$ in folded configuration during upstroke.}
  \label{fig:quad_folded}
\end{figure}

For the numerical example presented in this paper, we chose the base link length $a_0 = 80$~mm and the Bennett ratio $b = \frac{1}{2}$, which defines $a_1 = 2a_0 = 160$~mm as given in~\eqref{eq:bennett_cond}. The $x_i$ and $y_i$ coordinates of the quadrilateral points $p_i = [x_i, y_i, z_i] \in \mathbb{R}^3$, $i = 0\dots3$, are determined by $a_0$; however, the $z_i$ coordinates are free design parameters corresponding to the physical locations of link connections along their respective axes, and thus directly influencing the wing surface area. In this example, we define the points
\begin{equation}
    p_0 = 
    \begin{bmatrix}
        0 \\ 0 \\ 0
    \end{bmatrix}\;
    p_1 = 
    \begin{bmatrix}
        0 \\ 80 \\ z_1
    \end{bmatrix}\;
    p_2 = 
    \begin{bmatrix}
        160 \\ 80 \\ z_2
    \end{bmatrix}\;
    p_3 = 
    \begin{bmatrix}
        160 \\ 0 \\ z_3
    \end{bmatrix}, 
\end{equation}
and the corresponding folded configuration 
\begin{equation}
    p_0' = 
    \begin{bmatrix}
        0 \\ 0 \\ 0
    \end{bmatrix}\;
    p_1' = 
    \begin{bmatrix}
        0 \\ 80 \\ z_1
    \end{bmatrix}\;
    p_2' = 
    \begin{bmatrix}
        0 \\ 240 \\ z_2
    \end{bmatrix}\;
    p_3' = 
    \begin{bmatrix}
        0 \\ 160 \\ z_3
    \end{bmatrix}.
\end{equation}
Note that $p_0$ and $p_1$ form the base link and therefore remain equal to $p_0'$ and $p_1'$ in the folded configuration. With these values, planar four-bar linkages can be constructed, in which the parameters $z_{1\dots3}$ define the physical connection locations along the joint axes. 
In the numerical example, the values $z_1 = 10$~mm, $z_2 = 40$~mm, and $z_3 = -50$~mm were chosen. The schematic drawings of the expanded and folded configurations are shown in Figs.~\ref{fig:quad_expanded} and~\ref{fig:quad_folded}.

\begin{figure}[b]
  \centering
  \includegraphics[width=\columnwidth]{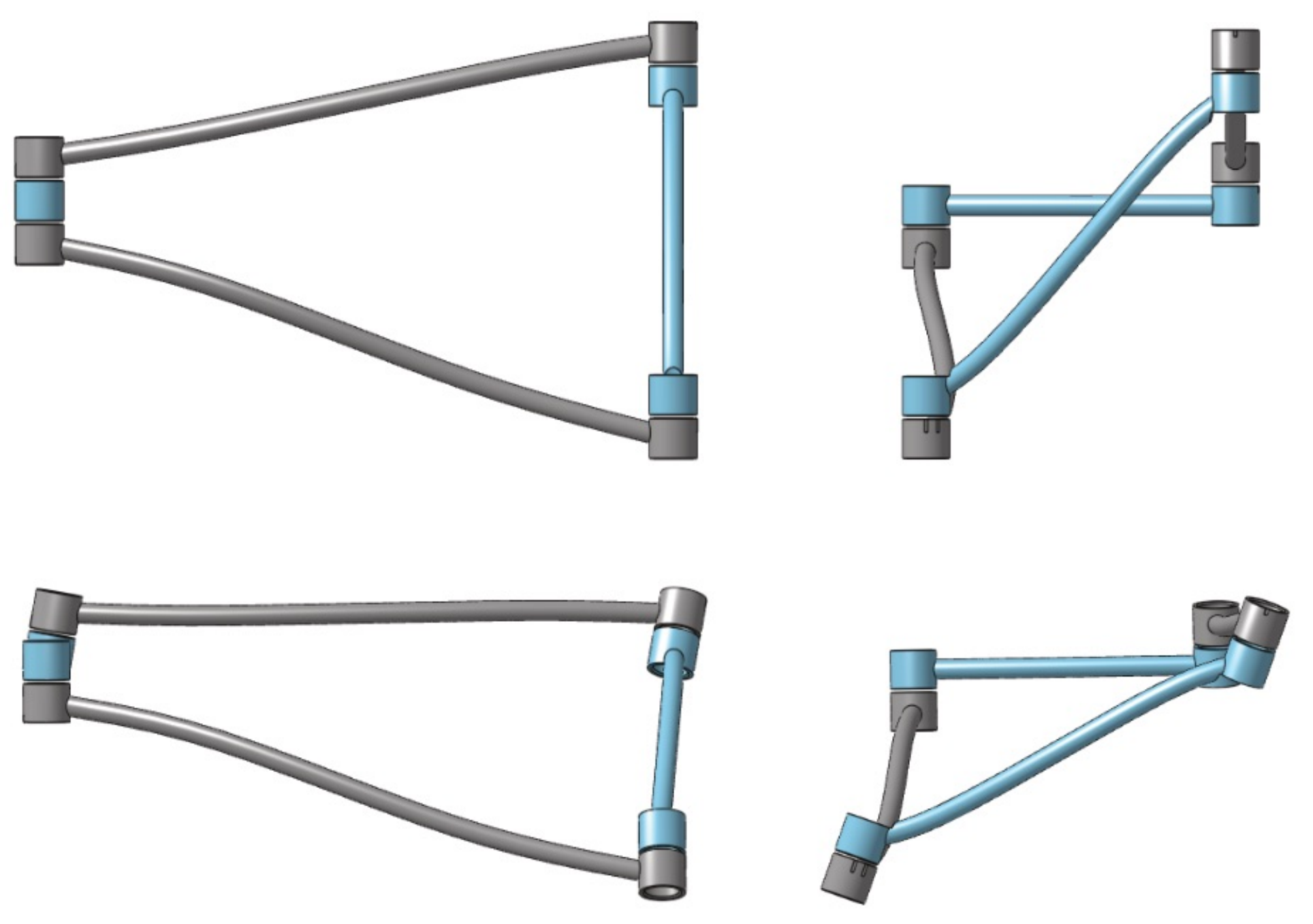}
  \caption{Front and right views comparison of planar linkage (above) and Bennett linkage with the twist given by $\alpha_0 = 10^{\circ}$ (bellow).}
  \label{fig:bennett_vs_planar}
\end{figure}

In a parametric computer-aided design (CAD) model, it is straightforward to introduce twist angles $\alpha_i$ between the joint axes to obtain the spatial Bennett linkage. For example, choosing $\alpha_0 = 10^{\circ}$ (yielding $\alpha_1 = 20.32^{\circ}$ according to~\eqref{eq:bennett_cond}) displaces points $p_2$ and $p_3$ as shown in Fig.~\ref{fig:bennett_vs_planar}, while keeping $p_0$ and $p_1$ fixed. This provides an additional degree of freedom in the design and is crucial for adjusting the wing-stroke angle. With the chosen twist angles, points $p_2$ and $p_3$ map to $p''_2 = [160,\ 90.2,\ 37.7]^T$ and $p''_3 = [162.1,\ -16.3,\ -17.9]^T$.

%This can be confirmed by using these points and calculating the bilinear patch area of the two quadrilaterals, which being $A^p = 11441$ mm\textsuperscript{2} of the planar linkage smaller than $A^b = 16267$ of the Bennett linkage. In Fig.~\ref{} this can also be observed visually.
In this way, a parametric model of a Bennett-type linkage was developed that adjusts the surface area spanned by the defining quadrilateral, and can serve as a basis for optimizing the desired surface shape between expanded and folded configurations in future work. The design parameters chosen for this numerical example will be further discussed in the context of embodied behavior in Sect.~\ref{sec:flapping_bennett}.

\subsection{Flapping Motion Mechanism}\label{sec:flapping_bennett}
%\subsection{Stroke-linkage for Spatial Sweep Motion}\label{sec:flapping_bennett}

This section describes the generation of flapping motion using two spatial Bennett linkages, the Stroke-linkage and the Folding-linkage. 

It is yet left to design the Stroke-linkage that will carry the Folding-linkage obtained in the previous section.
For the Stroke-linkage, the design strategy differs, as the motion will be generated from given poses rather than constructed from design parameters. The motion methodology through three poses \cite{Brunnthaler2005new} will be applied along with the MotionDesigner module \cite{Huczala2026ark} of the Python library Rational Linkages~\cite{Huczala2024ark}, which allows for interactive CAD construction.

In summary, the methodology of~\cite{Brunnthaler2005new} shows that three poses define a plane in dual quaternion space that intersects the Study quadric in a conic, representing a quadratic motion curve. These conics can be parameterized using two scalar parameters $\alpha$ and $\beta$, which must be chosen so that the curve lies entirely on the Study quadric. Substituting the parameterization into the quadric equation yields a quartic polynomial in $t$, which simplifies due to the parameterization structure and gives a linear system of equations that uniquely determines $\alpha$ and $\beta$. The result is a quadratic (Bennett) rational motion polynomial parameterized by a single parameter~$t$ (1-DoF).

Rational motions can be factorized into linear factors representing simple rotations, which correspond to the joint axes of a mechanism. For more details, see the rational motion factorization methodology of~\cite{Li2019factori}. The parameter $t$ then corresponds to the driving joint angle~\cite[Eq.~(9)]{Huczala2025directk}.

\begin{figure}[tb]
    \centering
    \includegraphics[width=\columnwidth, trim={0cm 1cm 0cm 0cm}, clip]{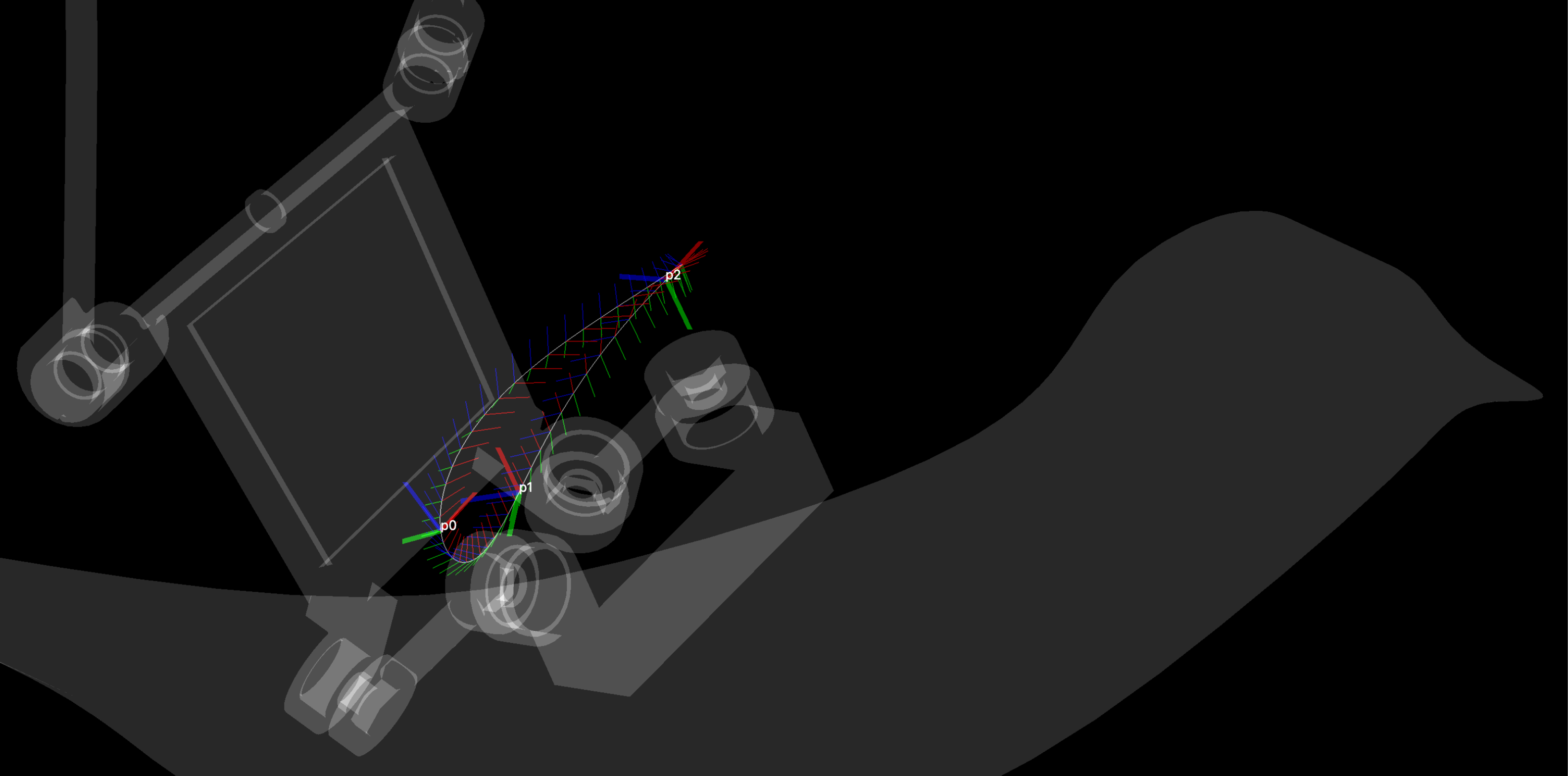}
    \caption{Continuous rational motion of the coupler interpolating three given poses; captured in the MotionDesigner module \cite{Huczala2026ark}.}
    \label{fig:motion_design}
\end{figure}

The motion and the resulting mechanism were obtained interactively using~\cite{Huczala2026ark}, as shown in Fig.~\ref{fig:motion_design}. The key design criterion was to define input poses $\mathbf{T}_0, \mathbf{T}_1, \mathbf{T}_2$ such that during one part of the motion -- when traveling from pose $\mathbf{T}_0$ to $\mathbf{T}_2$ via $\mathbf{T}_1$ -- the plane spanned by the $x$- and $y$-axes of the coupler sweeps forward, creating surface resistance. During the return of the closed-loop trajectory from $\mathbf{T}_2$ back to $\mathbf{T}_0$, the plane is tilted so that it cuts through the air. For the numerical example and wing design, the following poses were chosen, with $\mathbf{T}_0$ being the identity matrix, 
\begin{gather*}
    \mathbf{T}_1 = 
    \begin{pmatrix}
        0.592 & -0.103 & -0.799 & 0.030 \\
        0.571 & 0.753 & 0.326 & 0.018 \\
        0.569 & -0.649 & 0.505 & -0.012 \\
        0 & 0 & 0 & 1
    \end{pmatrix}, \\ 
    \text{and }\mathbf{T}_2 = 
    \begin{pmatrix}
        1 & 0 & 0 & 0.065\\
        0 & 0.28 & 0.96 & 0\\
        0 & -0.96 & 0.28 & 0\\
        0 & 0 & 0 & 1\\
    \end{pmatrix}.
\end{gather*}
Inputting poses $\mathbf{T}_{0\dots2}$ as dual quaternions into the algorithm of~\cite{Brunnthaler2005new} yields the following rational motion polynomial
\begin{gather*}
C(t) = \begin{pmatrix}
t^2 + 0.0052\,t + 0.8087 \\
-0.0140\,t - 0.6065 \\
-0.8702\,t \\
0.4286\,t \\
-0.0197 \\
0.0004\,t - 0.0263 \\
-0.0136\,t \\
0.0184\,t
\end{pmatrix},
\end{gather*}
which can be further factorized~\cite{Li2019factori} to obtain the joint axes of the Bennett mechanism with DH parameters
\begin{gather*}
    a_0 = 32.5\text{ mm, } \alpha_0 = 91.5^{\circ}, \\
    a_1 = 54.2\text{ mm, } \alpha_1 = 143.1^{\circ}.
\end{gather*}
Note that the input values above have been rounded for readability; however, the algorithms require high numerical precision. Therefore, it is advisable to orthogonalize the matrices $\mathbf{T}_{0\dots2}$ prior to use.
Finally, we can obtain the Stroke-linkage actuated by a motor mounted on the body frame of the bird-like structure, which provides the desired spatial stroke motion.

\begin{figure}[tb]
    \centering
    \includegraphics[width=\linewidth]{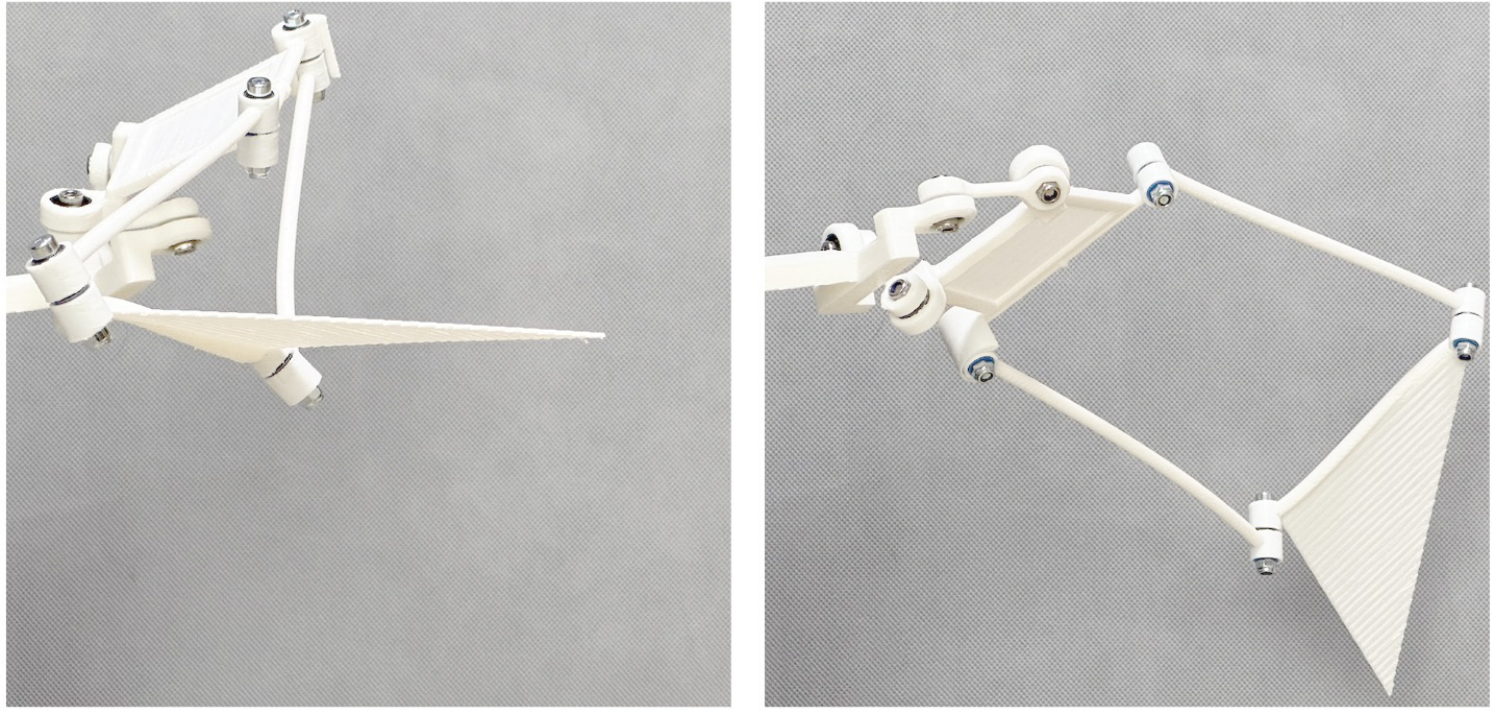}
    \includegraphics[width=\linewidth]{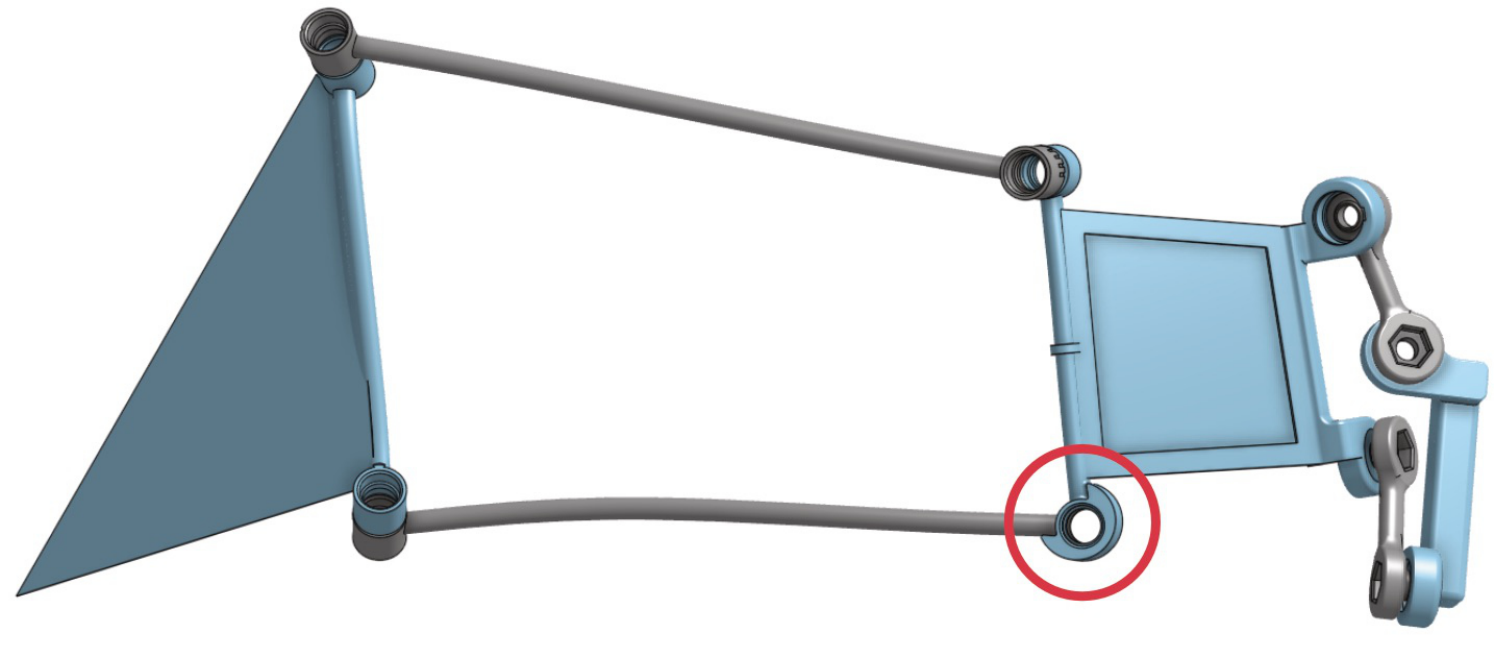}
    \caption{CAD model of the flapping wing (bellow) with physical joint limits (red circle); 3D printed prototype in its folded (top left) and extended (top right) configurations.}
    \label{fig:wing-cad}
\end{figure}

\addtolength{\textheight}{-0.5cm}

\section{Wing Prototype with Bennett Linkages}\label{sec:Prototype}

The Folding-linkage will automatically fold and extend based on the motion dynamics provided by the Stroke-linkage and the surrounding airflow. This embodied degree of freedom remains unactuated; however, to achieve the desired function, physical stops were placed on one of the joints of the Folding-linkage, as shown by the red circle in Fig.~\ref{fig:wing-cad}, constraining its rotation to a range of $75^{\circ}$ in both directions.

As shown in Fig.~\ref{fig:wing-cad}, the design parameters were chosen to produce the following behavior. When the wing is extended, the airflow acts on the bottom surface of the wing to lift the bird-like robot. In the folded configuration, the wing folds such that the airflow acts from the opposite side (on that top surface), keeping it folded during the upstroke motion.

An experimental design and assembly of a prototype of the wing was constructed and 3D printed in polylactide (PLA) material. The joints were assembled using ball bearings and M4 screws. The fully assembled wing weighs 87~grams. It is shown in Fig.~\ref{fig:wing-cad}.

A video demonstrating the kinematic design and behavior of the linkages can be found in the supplementary material. The video ends with real-time footage of a laboratory prototype test and is also available online in~\cite{zenodo_suppl}.

\section{CONCLUSIONS}\label{sec:conclusion}

This paper demonstrated the potential of spatial single-loop mechanisms for bio-inspired robotics applications. As a case study, a flapping wing mechanism was designed through kinematic synthesis. The proposed design consists of two coupled Bennett four-bar linkages that deliver spatial sweep motion and passive folding using only a single actuator. A physical prototype validated the expected kinematic behavior.

It must be noted that kinematic design and analysis alone do not guarantee aerodynamic performance. Therefore, future work will focus on dynamics modeling for flight simulation and aerodynamic shape optimization of the wing design parameters and a membrane. The wing membrane is currently under development; origami-like structures, as proposed in the design of the beetle wing in~\cite{Phan2020mechani}, present a promising and feasible approach for this application. For dynamics modeling, flexible multibody model \cite{Gerstmayr2023exudyna} will be employed as it is a suitable way to simulate overconstrained mechanisms. The available simulation tools allow for the inclusion of analytical aerodynamic models, such as the quasi-steady aerodynamic model based on strip theory~\cite{Kim2011improve}. Through the integration of these methodologies, we aim to develop a comprehensive wing modeling framework with design parameters that can be tuned to meet specific flight performance requirements.
Finally, full assembly of the bird-inspired robotic platform and flight testing are planned to validate the developed model and demonstrate the feasibility of the proposed methodology.

Beyond the presented application, the proposed design methodology offers a broader perspective. Spatial single-loop mechanisms can be integrated into a range of bio-inspired designs -- including robotic arm and leg structures, which often rely on planar pantograph mechanisms, as well as underwater robotic fins. Additionally, there exists a class of 1-DoF six-bar linkages that can perform even more complicated trajectories \cite{Hegedus2015fourpos}, however, the design methodology is similar. 
The fundamental transition from planar to spatial mechanism architectures may unlock new motion capabilities and open new directions in bio-inspired robot design.

 % This command serves to balance the column lengths
                                  % on the last page of the document manually. It shortens
                                  % the textheight of the last page by a suitable amount.
                                  % This command does not take effect until the next page
                                  % so it should come on the page before the last. Make
                                  % sure that you do not shorten the textheight too much.

%%%%%%%%%%%%%%%%%%%%%%%%%%%%%%%%%%%%%%%%%%%%%%%%%%%%%%%%%%%%%%%%%%%%%%%%%%%%%%%%

%%%%%%%%%%%%%%%%%%%%%%%%%%%%%%%%%%%%%%%%%%%%%%%%%%%%%%%%%%%%%%%%%%%%%%%%%%%%%%%%

%%%%%%%%%%%%%%%%%%%%%%%%%%%%%%%%%%%%%%%%%%%%%%%%%%%%%%%%%%%%%%%%%%%%%%%%%%%%%%%%
% \section*{APPENDIX}

% Appendixes should appear before the acknowledgment.

% \section*{ACKNOWLEDGMENT}

%%%%%%%%%%%%%%%%%%%%%%%%%%%%%%%%%%%%%%%%%%%%%%%%%%%%%%%%%%%%%%%%%%%%%%%%%%%%%%%%

%References are important to the reader; therefore, each citation must be complete and correct. If at all possible, references should be commonly available publications.

\bibliographystyle{IEEEtran}
\bibliography{IEEEabrv,references}

\end{document}